\documentclass[runningheads]{llncs}
\usepackage{graphicx}
\usepackage{tikz}
\usepackage{comment}
\usepackage{amsmath,amssymb} % define this before the line numbering.
\usepackage{color}

\usepackage[accsupp]{axessibility}  % Improves PDF readability for those with disabilities.

\usepackage{graphicx}
\usepackage{amsmath}
\usepackage{amssymb}
\usepackage{booktabs}
\usepackage{times}
\usepackage{epsfig}
\usepackage{mathtools}
\usepackage{multirow}
\usepackage{enumitem}
\usepackage{caption}
\usepackage{subcaption}
\usepackage{array}
\usepackage{colortbl}
\usepackage{bbm}
\usepackage{multicol}
\usepackage{makecell}
\usepackage{xcolor}
\usepackage{xspace}
\usepackage{float}
\usepackage{color}
\usepackage{pifont}
\usepackage{marvosym}
\usepackage{appendix}
\usepackage{wrapfig}
\usepackage[pagebackref,breaklinks,colorlinks]{hyperref}

\usepackage[capitalize]{cleveref}
\crefname{section}{Sec.}{Secs.}
\Crefname{section}{Section}{Sections}
\Crefname{table}{Table}{Tables}
\crefname{table}{Tab.}{Tabs.}

\def\vs{\emph{vs.}}
\def\ie{\emph{i.e.}}
\def\eg{\emph{e.g.}}

\def\etal{{\em et al.~}}

\newlength\savedwidth
\newcommand\whline{\noalign{\global\savedwidth\arrayrulewidth\global\arrayrulewidth 0.8pt}\hline\noalign{\global\arrayrulewidth\savedwidth}}
\definecolor{mygray}{gray}{.92}
\newcommand{\tabincell}[2]{\begin{tabular}{@{}#1@{}}#2\end{tabular}}

\newcommand\blfootnote[1]{%
\begingroup
\renewcommand\thefootnote{}\footnote{#1}%
\addtocounter{footnote}{-1}%
\endgroup
}

\begin{document}
% \linenumbers
\pagestyle{headings}
\mainmatter
\def\ECCVSubNumber{4849}  % Insert your submission number here

\title{VL-LTR: Learning Class-wise Visual-Linguistic Representation for Long-Tailed Visual Recognition} % Replace with your title

\begin{comment}
\titlerunning{ECCV-22 submission ID \ECCVSubNumber} 
\authorrunning{ECCV-22 submission ID \ECCVSubNumber} 
\author{Anonymous ECCV submission}
\institute{Paper ID \ECCVSubNumber}
\end{comment}
%******************

%\begin{comment}
\titlerunning{VL-LTR}
\author{Changyao Tian$^{1*\dag}$, 
    Wenhai Wang$^{3*}$,
    Xizhou Zhu$^{2*}$,
    Jifeng Dai$^{2}$\textsuperscript{\Letter},
    Yu Qiao$^{3}$ \\
    $^1$Chinese University of Hong Kong~~~~
    $^2$SenseTime~~~~
    $^3$Shanghai AI Laboratory
    {\tt\small tcyhost@buaa.edu.cn}~~~~
    {\tt\small \{wangwenhai, qiaoyu\}@pjlab.org.cn}\\
    {\tt\small \{zhuwalter, daijifeng\}@sensetime.com}~~~~
}
\authorrunning{Tian et al.}
\institute{}
%\end{comment}
%******************
\maketitle

\begin{abstract}
% Deep learning-based models encounter challenges when processing long-tailed data in the real world. Existing solutions usually employ some balancing strategies or transfer learning to deal with the class imbalance problem, based on the image modality. In this work, we present a visual-linguistic long-tailed recognition framework, termed VL-LTR, and conduct empirical studies on the benefits of introducing text modality for long-tailed recognition (LTR). 
Recently, computer vision foundation models such as CLIP and ALI-GN, have shown impressive generalization capabilities on various downstream tasks. But their abilities to deal with the long-tailed data still remain to be proved. In this work, we present a novel framework based on pre-trained visual-linguistic models for long-tailed recognition (LTR), termed VL-LTR, and conduct empirical studies on the benefits of introducing text modality for long-tailed recognition tasks.
Compared to existing approaches, the proposed VL-LTR has the following merits. 
(1) Our method can not only learn visual representation from images but also learn corresponding linguistic representation from noisy class-level text descriptions collected from the Internet; 
(2) Our method can effectively use the learned visual-linguistic representation to improve the visual recognition performance, especially for classes with fewer image samples. We also conduct extensive experiments and set the new state-of-the-art performance on widely-used LTR benchmarks. Notably, our method achieves 77.2\% overall accuracy on ImageNet-LT, which significantly outperforms the previous best method by over 17 points, and is close to the prevailing performance training on the full ImageNet. Code is available at \url{https://github.com/ChangyaoTian/VL-LTR}.

\blfootnote{$*$ Authors contributed equally. \Letter~Corresponding author. }
\blfootnote{$\dag$ The work is done when Changyao Tian is an intern at SenseTime Research.}
\end{abstract}

\section{Introduction}

\label{sec:intro}

\begin{figure}[t]
        % \vspace{8pt}
		\centering
		\setlength{\fboxrule}{0pt}
		\fbox{\includegraphics[width=0.95\textwidth]{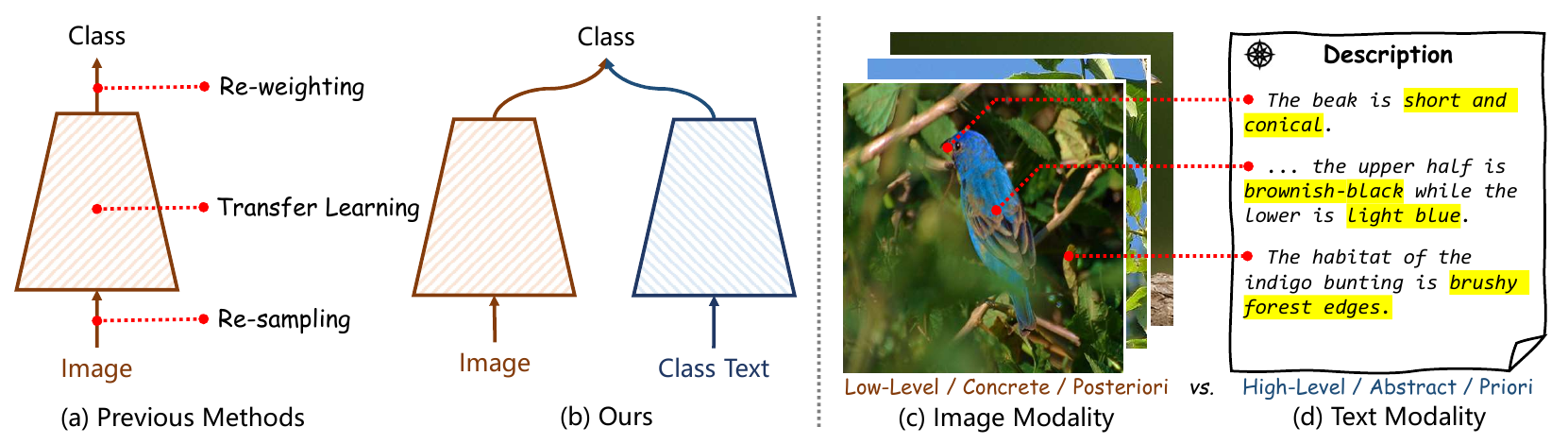}}
		% \vspace{4pt}
		\caption{
		\textbf{Comparison of different long-tailed recognition (LTR) frameworks and different modalities. } 
		(a) Previous LTR methods~\cite{chawla2002smote,chu2020feature,khan2017cost,cui2019class,wang2017learning,zhong2019unequal} mainly focus on the class imbalance problem on the image modality, while (b) our method addresses the LTR task by combining the advantages of image and text modalities.
		(c) and (d) give intuitive explanations for the correlations and differences between the image and text modalities.
		}
		\label{fig:moti}
		% \vspace{-20pt}
\end{figure}

Real-world data always presents a long-tailed distribution, where only a few head classes encompass most of the data, and most tail classes have very few samples. Such phenomenon is not conducive to the practical application of deep-learning based models.
Because of this, a number of works have emerged and tried to alleviate the class imbalance problem from different aspects, such as re-sampling the training data~\cite{chawla2002smote,chu2020feature,shen2016relay}, re-weighting the loss functions~\cite{khan2017cost,cui2019class,lin2017focal}, or employing transfer learning methods~\cite{wang2017learning,zhong2019unequal,liu2019large} (see Figure \ref{fig:moti} (a)).
Despite their great contributions, most of these works still restrict themselves to only relying on the image modality for solving this problem. 

As illustrated in Figure \ref{fig:moti} (c)(d), there are some inner connections between images and text descriptions of the same class, especially when it comes to some visual concepts and attributes.
However, different from the image modality that usually presents concrete low-level features (\eg, shape, color, texture) of the object or scene, the text modality typically contains much high-level and abstract information.
Furthermore, text descriptions are prior knowledge that can be summarized by experts, which could be useful when there are no sufficient images to learn general class-wise representation for recognition.

Although there have been some visual-linguistic approachs~\cite{he2017fine,zhuang2020wildfish++,mu2019shaping} for visual recognition, their performance is still not satisfactory, due to the gap between image and text representation and the lack of robustness to noisy text.
%
% Recently, the rise of visual-linguistic pre-training~\cite{clip,su2019vl,jia2021scaling} has provided an effective way to learn powerful representation that can connect the image and text modalities.
% \tcy{
Recently, the rise of visual-linguistic foundation models~\cite{clip,su2019vl,jia2021scaling} has provided an effective way to learn powerful representation that can connect the image and text modalities.
% }
%
Motivated by this, we present a visual-linguistic framework for long-tailed recognition, termed VL-LTR, which can utilize the advantages of both visual and linguistic representation for visual recognition tasks as shown in Figure \ref{fig:moti} (b).
Our method mainly consists of two key components, which are (1) a class-wise visual-linguistic pre-training (CVLP) framework for linking images and text descriptions at the class level, and (2) a language-guided recognition (LGR) head designed to perform long-tailed recognition according to the learned visual-linguistic representation.

Overall, the proposed VL-LTR possesses the following merits. (1) Compared to
the visual-linguistic pre-training \cite{cui2021parametric,zhang2021test,cui2021reslt,wang2020long,kang2019decoupling}, our method can learn visual-linguistic
representation at the class level, and take the advantages of class-wise linguistic representation to improve visual recognition performance, especially in the long-tailed scenario;
(2) Compared to previous visual-linguistic classifiers~\cite{he2017fine,zhuang2020wildfish++,mu2019shaping}, our method can not only effectively bridge the gap between visual and linguistic representation, but also be more flexible and robust to noisy text descriptions.

\begin{wrapfigure}{r}{0.59\textwidth}
    % \vspace{-6pt}
    \begin{center}
    \setlength{\fboxrule}{0pt}
    		\fbox{\includegraphics[width=0.5\textwidth]{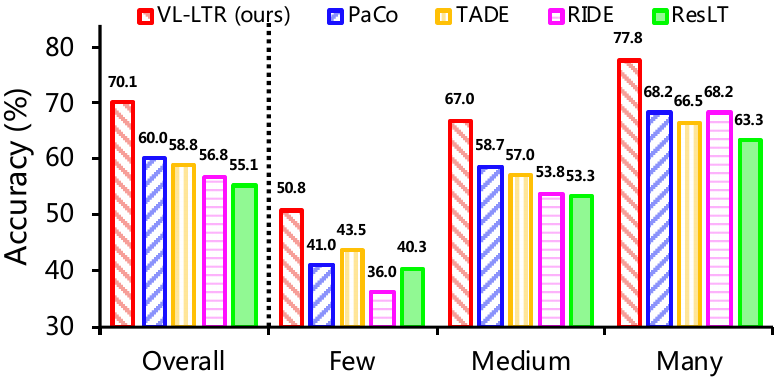}}
    	\end{center}
    	\captionsetup{font={scriptsize}}
    	% \vspace{-20pt}
    \caption{
    \textbf{Performance comparison on ImageNet-LT~\cite{liu2019large}.} Our VL-LTR (ResNet-50~\cite{he2016deep}) significantly outperforms prior arts, including PaCo~\cite{cui2021parametric}, TADE~\cite{zhang2021test}, RIDE (4 Experts)~\cite{wang2020long}, and ResLT~\cite{cui2021reslt}, which use heavier ResNeXt-50/101~\cite{xie2017aggregated} as backbone.
    }
    \label{fig:res_intro}
    % \vspace{-15pt}
\end{wrapfigure}

To verify the effectiveness of our method, we conduct extensive experiments on three challenging long-tailed recognition (LTR) benchmarks, including ImageNet-LT~\cite{liu2019large}, Places-LT~\cite{liu2019large}, and iNaturalist 2018~\cite{van2018inaturalist}. 
As shown in Figure \ref{fig:res_intro}, using ResNet-50~\cite{he2016deep} as backbone, our method achieves an overall accuracy of 70.1\% on ImageNet-LT, which is 10.1 points higher than the previous best method PaCo~\cite{cui2021parametric} (ResNeXt-101~\cite{xie2017aggregated}).
For tail classes, the medium and few-shot accuracy of our method reaches 67.0\% and 50.8\% respectively, which significantly outperform that of the prior arts~\cite{cui2021parametric,zhang2021test,cui2021reslt,wang2020long} as well.

% \begin{wrapfigure}[r]
%  		\centering
% 		\setlength{\fboxrule}{0pt}
% 		\fbox{\includegraphics[width=0.44\textwidth]{./figure/res_intro1.pdf}}
% 		% \vspace{-8pt}
% 		\caption{
% 		\textbf{Performance comparison on ImageNet-LT~\cite{liu2019large}.} Our VL-LTR (ResNet-50~\cite{he2016deep}) significantly outperforms prior arts, including PaCo~\cite{cui2021parametric}, TADE~\cite{zhang2021test}, RIDE (4 Experts)~\cite{wang2020long}, and ResLT~\cite{cui2021reslt}, which use heavier ResNeXt-50/101~\cite{xie2017aggregated} as backbone.
% 		}
% 		\label{fig:res_intro}
% 		\vspace{-12pt}
% \end{wrapfigure}

In summary, our main contributions are three-fold.

(1) We provide a detailed analysis on the connection and differences between image and text modalities, and point out that class descriptions can serve as a supplement to images, which is conducive to long-tailed visual recognition.

(2) We present a new visual-linguistic framework for long-tailed visual recognition (VL-LTR), which contains two tailored components, including a class-wise text-image pre-training (CVLP) to bridge the class-level images and text descriptions, and a language-guided recognition (LGR) head to perform classification based on the learned visual-linguistic representation.

(3) The proposed VL-LTR has achieved state-of-the-art performance on prevailing ImageNet-LT, Places-LT, and iNaturalist 2018 datasets. Notably, our method gets the best overall accuracy of 77.2\% on ImageNet-LT, outperforming the old record by 17.2 points, and even approaching the performance training on the full ImageNet~\cite{deng2009imagenet}.

\section{Related Work}

\subsection{Long-Tailed Visual Recognition}
Class re-balanced strategy~\cite{han2005borderline,khan2017cost,cao2019learning,drummond2003c4,wang2017learning,jamal2020rethinking}
has been comprehensively studied for long-tailed visual recognition.
One type of the class re-balanced strategy is Data Re-sampling~\cite{drummond2003c4,buda2018systematic,chawla2002smote,han2005borderline,chu2020feature,shen2016relay}, which generates class-balanced data by adjusting the sampling rate of tail classes and head classes, yet they might take the risk of over-fitting on data-scarced classes.
Besides that, some recent methods~\cite{chu2020feature,kim2020m2m} augment tail class samples with head classes ones, to alleviate the over-fitting problem. 
Another kind of class re-balanced strategy is to design re-weighting loss functions, where tail classes would be emphasized by using large weights or margins~\cite{cui2019class,khan2017cost,huang2016learning,wang2017learning,cao2019learning}, or ignoring negative gradients for tail classes~\cite{tan2020equalization}.

In addition, researchers also address the long-tailed recognition task from the aspect of transfer learning~\cite{liu2019large,zhu2020inflated,yin2019feature,kang2019decoupling,zhou2020bbn}.
Liu \etal~\cite{liu2019large} and Zhu \etal~\cite{zhu2020inflated} transfer knowledge from head classes's features to
tail classes by maintaining memory bank and modeling intra-class variance, respectively.
After that, Samuel \etal~\cite{samuel2021generalized} proposes a late-fusion framework for long-tail learning with class descriptors.
Some decoupling methods~\cite{zhou2020bbn,kang2019decoupling} also can be regarded as transferring head classes frozen feature to tail classes when fine-tuning classifiers. 
Recently, some studies~\cite{kang2020exploring,wang2021contrastive,cui2021parametric,samuel2021distributional,zhang2021test} also transfer the representation learned by contrastive learning or self-supervised learning for long-tailed problems.

The aforementioned methods mainly focus on addressing the class imbalance problem based on image modalities, while rarely exploring the possibility of integrating text modalities on this problem.

\subsection{Visual-linguistic Model}
In this section, we mainly discuss visual-linguistic pre-training and classification related to our work.

Visual-linguistic pre-training~\cite{lu2019vilbert,tan2019lxmert,chen2019uniter,li2020oscar,zhang2021vinvl,li2020closer,gan2020large,li2020hero,lu202012,clip} have achieved great success on a number of downstream vision tasks. 
Zhang \etal~\cite{zhang2021vinvl} show the importance of visual features in visual-linguistic pre-training and obtain more strong visual representations from large object detectors.
Li \etal~\cite{li2020oscar} find that a larger transformer visual-linguistic model can learn more powerful representation from a larger visual-linguistic corpus.
In addition, Huang \etal~\cite{huang2020pixel,su2019vl} proposed a visual-linguistic pre-training model by extracting patch features from the convolutional layers without the proposal computation.
Recently, CLIP~\cite{clip} and ALIGN~\cite{jia2021scaling} learns powerful visual-linguistic representation via contrastive learning on large-scale image-text pairs.

Prior to these works, there have been some visual-linguistic approaches~\cite{he2017fine,zhuang2020wildfish++,mu2019shaping} designed for tasks related to image classification.
He \etal~\cite{he2017fine} propose a two-stream model, which directly combines visual and linguistic representation for fine-grained image classification.
Mu \etal~\cite{mu2019shaping} present a few-shot visual recognition model that is regularized with text descriptions during training.
Similar to He \etal~\cite{he2017fine}, Zhuang \etal~\cite{zhuang2020wildfish++} design a multi-modal model  for automatic fish classification, with a CNN encoder for images and a RNN encoder for class text. However, these methods (1) cannot effectively model the connection between images and text, leading to a considerable gap between visual and linguistic representation; and (2) require high-quality text annotations, which is usually expensive and thus limits their practical application.

\section{Methodology}

\subsection{Overall Architecture}
\label{sec:overall}

\begin{figure*}
		\centering
		\setlength{\fboxrule}{0pt}
		\fbox{\includegraphics[width=0.95\textwidth]{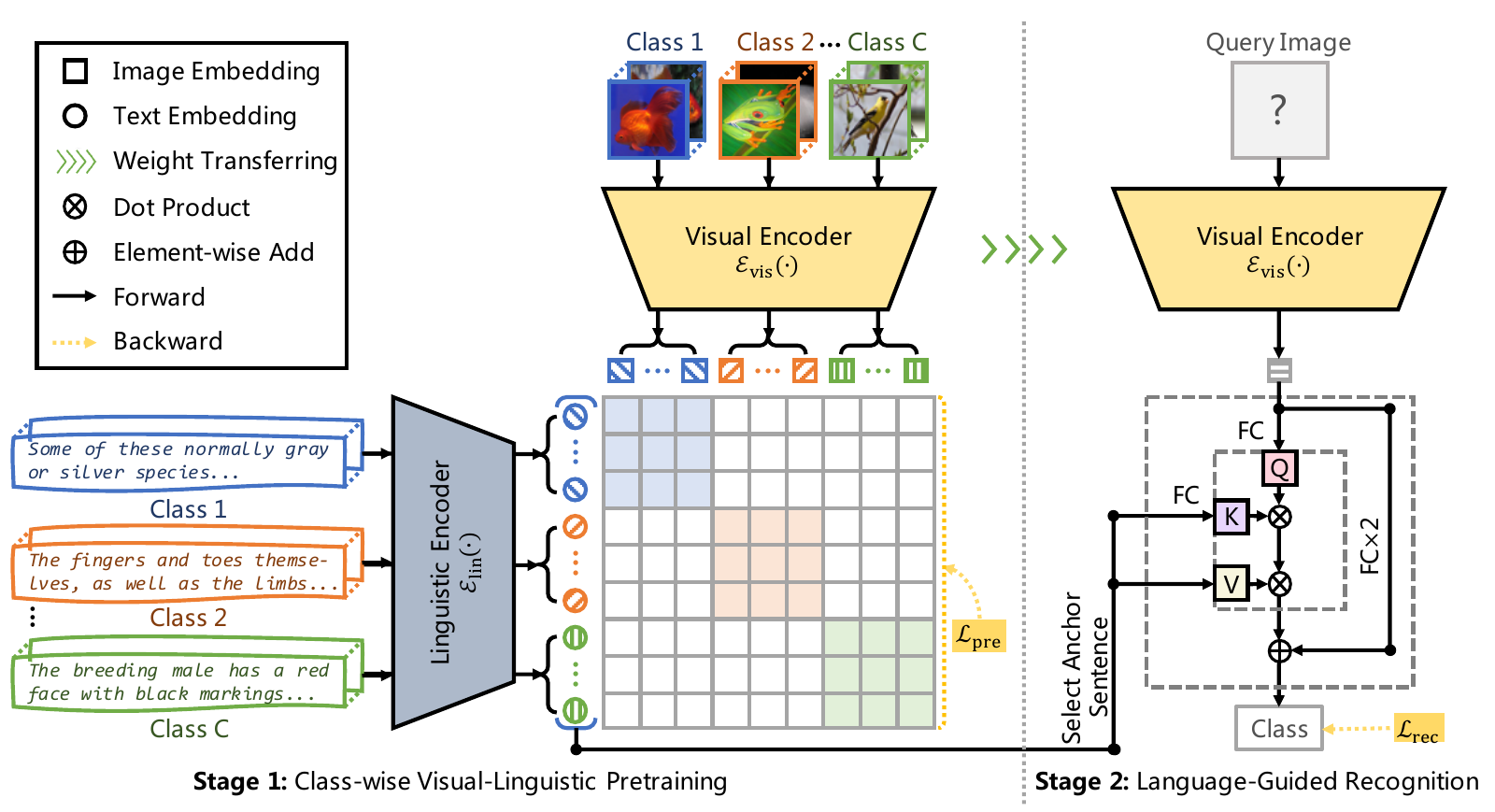}}
		% \vspace{-8pt}
		\caption{
		\textbf{Overall architecture of VL-LTR.}
		The entire model has two stages.
		In the first stage, class-wise visual-linguistic pre-training (CVLP) takes both the images and text of each class as inputs, learning to connect the representation of the two modalities through class-wise contrastive learning.
		In the second stage, the language-guided recognition (LGR) head uses the learned visual-linguistic representation to perform image classification.
		}
		\label{fig:arch}
		% \vspace{-16pt}
\end{figure*}

In order to make effective use of the linguistic modality in the visual recognition task, we propose a two-stage framework, as depicted in Figure \ref{fig:arch}. 
(1) The first stage is class-wise visual-linguistic pre-training (CVLP), which is used to link the images and text descriptions of the same class via contrastive learning.
(2) In the second stage, a language-guided recognition (LGR) head is designed to collect the overall linguistic representation of each class to guide the image recognition.
As a result, the proposed VL-LTR is able to combine the advantages of visual and linguistic representation and achieve impressive long-tailed recognition performance.

When training VL-LTR models, 
we first pre-train the visual and linguistic encoders by class-wise visual-linguistic contrastive learning and the pre-training loss $\mathcal{L}_\text{pre}$. 
During pre-training, an image and a sentence from the same class would be regarded as a positive pair, and otherwise is a negative pair.
After pre-training, the weights of the linguistic encoder are frozen, and the anchor sentences of each class are then selected by filtering out the low-scored sentences in text descriptions. 
The visual encoder and LGR head are fine-tuned by the recognition loss $\mathcal{L}_\text{rec}$. 
Details of the aforementioned loss functions will be introduced in the later sections.

In the inference phase, given a query image and pre-populated text embeddings of anchor sentences, we first feed the image to the visual encoder and obtain an image embedding.
Then, the image embedding passes through the LGR head and is categorized into a class according to the image embedding itself as well as the text embeddings of anchor sentences. 

\subsection{Class-wise Visual-Linguistic Pre-training}

The goal of this stage is to learn the visual-linguistic representation of images and text descriptions at the class level.
To this end,
we design a class-wise visual-linguistic pre-training (CVLP) framework. % inspired by [*].
Unlike previous works~\cite{clip,jia2021scaling} that use instance-wise image-text pairs for pre-training, 
our framework is expected to fuse the class-wise linguistic information into the visual space.

During pre-training, as shown in Figure \ref{fig:arch}, we first randomly sample a batch of images $\mathcal{I}=\left\{I_i \right\}_{i=1}^N$, and the corresponding text sentences $\mathcal{T}=\left\{T_i \right\}_{i=1}^N$, where $N$ denotes the batch size.
Then, the images $\mathcal{I}$ and texts $\mathcal{T}$ are fed to the visual encoder $\mathcal{E}_\text{vis}(\cdot)$ and linguistic encoder $\mathcal{E}_\text{lin}(\cdot)$ respectively, yielding image and text embeddings as Eqn. \ref{eqn:enc}:
\begin{align}
E^I_i = \mathcal{E}_\text{vis}(I_i),\ \ \ \
E^T_i = \mathcal{E}_\text{lin}(T_i),
\label{eqn:enc}
\end{align}
where both $E^I_i$ and $E^T_i$ are of $D$ dimensions. After that, a class-wise contrastive learning (CCL) loss is used to optimize the visual and linguistic encoders.
Let us denote the cosine similarity of $E^I_i$ and $E^T_j$ as $S_{i,j}$, and then the CCL loss can be formulated as:
\begin{equation}
\begin{aligned}
\mathcal{L}_\text{ccl} =&
\mathcal{L}_\text{vis} + \mathcal{L}_\text{lin}\\
=& -\frac{1}{|\mathcal{T}^+_i|} \sum_{T_j \in \mathcal{T}^+_i}  \text{log} \frac{\text{exp}(S_{i,j} / \tau)}{\sum_{T_k\in \mathcal{T}} \text{exp}(S_{i,k} / \tau)} \\
 &- \frac{1}{|\mathcal{I}_i^+|} \sum_{I_j \in \mathcal{I}_i^+}  \text{log} \frac{\text{exp}(S_{j,i} / \tau)}{\sum_{I_k \in \mathcal{I}} \text{exp}(S_{k,i} / \tau)},
 \label{eqn:l_ccl}
\end{aligned}
\end{equation}
where $\mathcal{L}_\text{vis}$ and $\mathcal{L}_\text{lin}$ denote the loss of visual and linguistic side respectively, while $\mathcal{T}^+_i$ denotes a subset of $\mathcal{T}$, in which each text shares the same class with the image $I_i$. Correspondingly, all images in $\mathcal{I}^+_i$ share the same class with the text $T_i$. $\tau$ is a learnable parameter with an initial value of 0.07.

In addition to CCL, we also distill the knowledge from the CLIP~\cite{clip} pre-trained model, to reduce the risk of over-fitting caused by limited text corpus in the pre-training stage.
The distillation loss $\mathcal{L}_\text{dis}$ can be written as Eqn. \ref{eqn:l_dis}:
\begin{equation}
\begin{aligned}
    \mathcal{L}_\text{dis}\!=\!-\frac{\text{exp}(S_{i,i}' / \tau)}{\sum_{T_j \in \mathcal{T}} \text{exp}(S_{i,j}' / \tau)} \text{log} \frac{\text{exp}(S_{i,i} / \tau)}{\sum_{T_k \in \mathcal{T}} \text{exp}(S_{i,k} / \tau)} \\
    -\frac{\text{exp}(S_{i,i}' / \tau)}{\sum_{I_j \in \mathcal{I}} \text{exp}(S_{j,i}' / \tau)} \text{log} \frac{\text{exp}(S_{i,i} / \tau)}{\sum_{I_k \in \mathcal{I}} \text{exp}(S_{k,i} / \tau)}.
\label{eqn:l_dis}
\end{aligned}
\end{equation}
Here, $S'$ is the cosine similarity matrix produced by the frozen CLIP model.

Our pre-training framework has two merits as follows:
(1)
It is convenient to add new training samples for image or text modality in our framework,
since the image and text description for a specific class is independent of each other, which greatly reduces the cost of data collection;
(2) The text description of each image sample is different in each iteration, which serves as an additional regularization to prevent the model from learning some fixed trivial correlation within a certain image-text pair, and thus our framework is robust to the noisy text from the Internet.

\subsection{Language-Guided Recognition}

In this stage, we design (1) an anchor sentence selection strategy to filter out noise texts, and (2) a language-guided recognition head to effectively use visual and linguistic representation learned in the pre-training stage.

\noindent\textbf{Anchor Sentence Selection.} Most text descriptions in our corpus are crawled from the Internet, which are noisy and might degrade the recognition performance. 
To address this problem, we propose an anchor sentence selection (AnSS) strategy to find the most discriminative sentences for each class. 
Specifically, we first construct a ``special'' image batch $I'$, which contains at most 50 images (if any) of each class.
Then, for each text sentence $T_i$, 
we score each sentence $T_i$ by computing the $\mathcal{L}_\text{lin}$ between the sentence and the image batch $I'$.
Finally, we select $M$ text sentences with the smallest $\mathcal{L}_\text{lin}$ as the anchor sentences for the follow-up visual recognition.

\noindent\textbf{Language-Guided Recognition Head.}
After obtaining the anchor sentences of each class, we design a language-guided recognition (LGR) head, to adjust the weights of these sentences based on the attention scores with the input image.
In this way, visual and linguistic features can be flexibly and dynamically combined according to the query image.

As shown in Figure \ref{fig:arch}, given an image embedding $E^I \in \mathbb{R}^D$, as well as the embeddings of all classes' anchor sentences $E^T \in \mathbb{R}^{C\times M\times D}$, where $C$ is the class number, and $M$ is the maximum number of sentences for each class. Then the LGR head can be formulated as:
\begin{align}
Q &= \text{Linear}(\text{LayerNorm}(E^I)), \\
K &= \text{Linear}(\text{LayerNorm}(E^T)),\ \ \ \ V = E^T, \\
G &= \sigma (\frac{Q K^\mathsf{T}}{\sqrt{D}}) V, \\
    P &= P^I + P^T = \sigma(\text{MLP}(E^I)) + 
    \sigma(\left<E^I, G\right>/\tau).
\end{align}
Here, $Q \in \mathbb{R}^D$, $K \in \mathbb{R}^{C\times M\times D}$, and $V \in \mathbb{R}^{C\times M\times D}$ are query, key and value of the attention operation.
$G\in \mathbb{R}^{C\times D}$ is the gather of the $M$ anchor sentence embeddings of each class.
$\sigma(\cdot)$ denotes Softmax function.
$\text{MLP}(\cdot)$ denotes two linear layers sandwich a ReLU in the middle.
$\left<E^I, G\right>$ is the cosine similarity of $E^I$ and $G$.
$P$ is the classification probability of the image $I_q$, $P^I$ and $P^T$ are the classification probabilities based on visual and linguistic representation, respectively.

\subsection{Loss Function}

As mentioned in Section \ref{sec:overall}, the training process of our method has two stages, namely pre-training and fine-tuning respectively.
In the pre-training stage, the visual encoder and linguistic encoder are jointly optimized by the CCL loss $\mathcal{L}_\text{ccl}$ and distillation loss $\mathcal{L}_\text{dis}$. So the overall pre-training loss can be written as:
\begin{equation}
\label{equ:img-loss}
    \mathcal{L}_\text{pre} = \lambda \mathcal{L}_\text{ccl} + (1 - \lambda) \mathcal{L}_\text{dis},
\end{equation}
where $\lambda \in [0,1]$ is a hyperparameter to balance $\mathcal{L}_\text{ccl}$ and $\mathcal{L}_\text{dis}$.

In the fine-tuning stage, after computing the classification probabilities $P^I$ and $P^T$, we simply calculate their corresponding CrossEntropy loss $\mathcal{L}_\text{CE}$ with the ground truth label $\mathbf{y}$ as Eqn. \ref{eqn:l_ce}:
\begin{equation}
    \mathcal{L}_\text{rec} = \mathcal{L}_\text{CE}(P^I, \mathbf{y}) + \mathcal{L}_\text{CE}(P^T, \mathbf{y}).
    \label{eqn:l_ce}
\end{equation}

\section{Experiments}

\subsection{Datasets}
We perform extensive experiments on three challenging long-tailed
visual recognition benchmarks, namely ImageNet-LT~\cite{liu2019large}, Places-LT~\cite{liu2019large}, and iNaturalist 2018~\cite{van2018inaturalist}.
Among these benchmarks, ImageNet-LT is constructed from ImageNet-2012~\cite{deng2009imagenet} by sampling a subset following the Pareto distribution with the power value $\alpha = 6$, which contains 1,000 classes.
The training set has 115.8K images, and the number of images per class ranges from 1,280 to 5 images. Both the validation set and the test set are balanced, containing 20K and 50K images respectively. We select the hyper-parameters on the validation set and report numerical results on the test set.
Similar to ImageNet-LT, Places-LT is a long-tailed version of the large-scale scene classification dataset Places~\cite{zhou2017places}. It consists of 62.5K images from 365 categories with class cardinality ranging from 5 to 4,980.
iNaturalist 2018 is a real-world, naturally long-tailed
dataset, which is composed of 8,142 fine-grained species.
The training set contains 437.5K images, and its imbalance
factor is equal to 500. We use the official validation set to
test our approach, which has 3 images per class.

We also collect the class-level text descriptions for the three datasets.
The text descriptions mainly come from Wikipedia\footnote{\url{https://en.wikipedia.org/}}, an open-source online encyclopedia that contains millions of articles for free. 
We first use the original class name as an initial query to get the best matching entry on Wikipedia.
After cleaning and filtering out some obviously irrelevant sections such as ``references'' or ``external links'' of these entries,
we split the left into sentences to form the original text candidate set for each class. 
Noting that some classes have relatively much fewer sentences than others, we also add 80 additional prompt sentences for each class to alleviate the data scarcity problem.
These sentences, which are in the form of `\texttt{a photo of a \{label\}}', are auto-generated based on the prompt templates provided in \cite{clip}.

\subsection{Evaluation Protocol} 
Following common practices~\cite{liu2019large,zhou2020bbn,cui2021parametric}, we evaluate our proposed models on the corresponding balanced validation/test set and report the overall top-1 accuracy. 
To diagnose the source of improvement, we also report the top-1 accuracy of the three subsets split by the number of training samples in each class, namely many-shot ($\ge$100 samples), medium-shot (20$\sim$100 samples), and few-shot ($\le$20 samples).

\subsection{Experiments on ImageNet-LT}
\label{sec:imnet_lt}

% \begin{figure}[t]
% 		\centering
% 		\setlength{\fboxrule}{0pt}
% 		\fbox{\includegraphics[width=0.46\textwidth]{./figure/range_acc.png}}
% 		\vspace{-8pt}
% 		\caption{
% 		\textbf{Absolute accuracy score of our method over the baseline using ViT-Base/16~\cite{dosovitskiy2020image} as the backbone on ImageNet-LT~\cite{liu2019large}.}
% 		Our method enjoys more performance gains on classes with fewer image samples.
% 		}
% 		\label{fig:range_acc}
% 		\vspace{-16pt}
% \end{figure}

\noindent\textbf{Settings.} To verify the effectiveness of our method, we conduct extensive experiments on ImageNet-LT~\cite{liu2019large}.
We use ResNet-50~\cite{he2016deep} or ViT-Base/16\cite{dosovitskiy2020image} as the visual encoder, and a 12-layer Transformer~\cite{radford2019language} as the linguistic encoder.
All models are optimized by AdamW~\cite{loshchilov2017decoupled} with a momentum of 0.9 and a weight decay of $5\times 10^{-2}$.
We use the same data augmentation as \cite{touvron2020training} (w/o distillation).
In the pre-training phase, the maximum length of text tokens is set to 77 (including \texttt{[SOS]} and \texttt{[EOS]} tokens), and the pre-trained weights of CLIP~\cite{clip} is loaded. The initial learning rate is set to $5\times 10^{-5}$ and decays following the cosine schedule~\cite{loshchilov2016sgdr}. During this phase, models are pre-trained for 50 epochs, with a mini-batch size of 256.
In the fine-tuning phase, we select 64 sentences for each class and fine-tune models with the mini-batch size of 128 for another 50 epochs. We set the initial learning rate to $1\times 10^{-3}$ and still decrease it with the cosine schedule.
In both stages, we adopt the input size of $224\times 224$ and the square-root data sampling strategy~\cite{mahajan2018exploring,mikolov2013distributed}  unless specifically mentioned.

\begin{wrapfigure}{r}{0.59\textwidth}
    % \vspace{-26pt}
    \begin{center}
    \setlength{\fboxrule}{0pt}
    		\fbox{\includegraphics[width=0.59\textwidth]{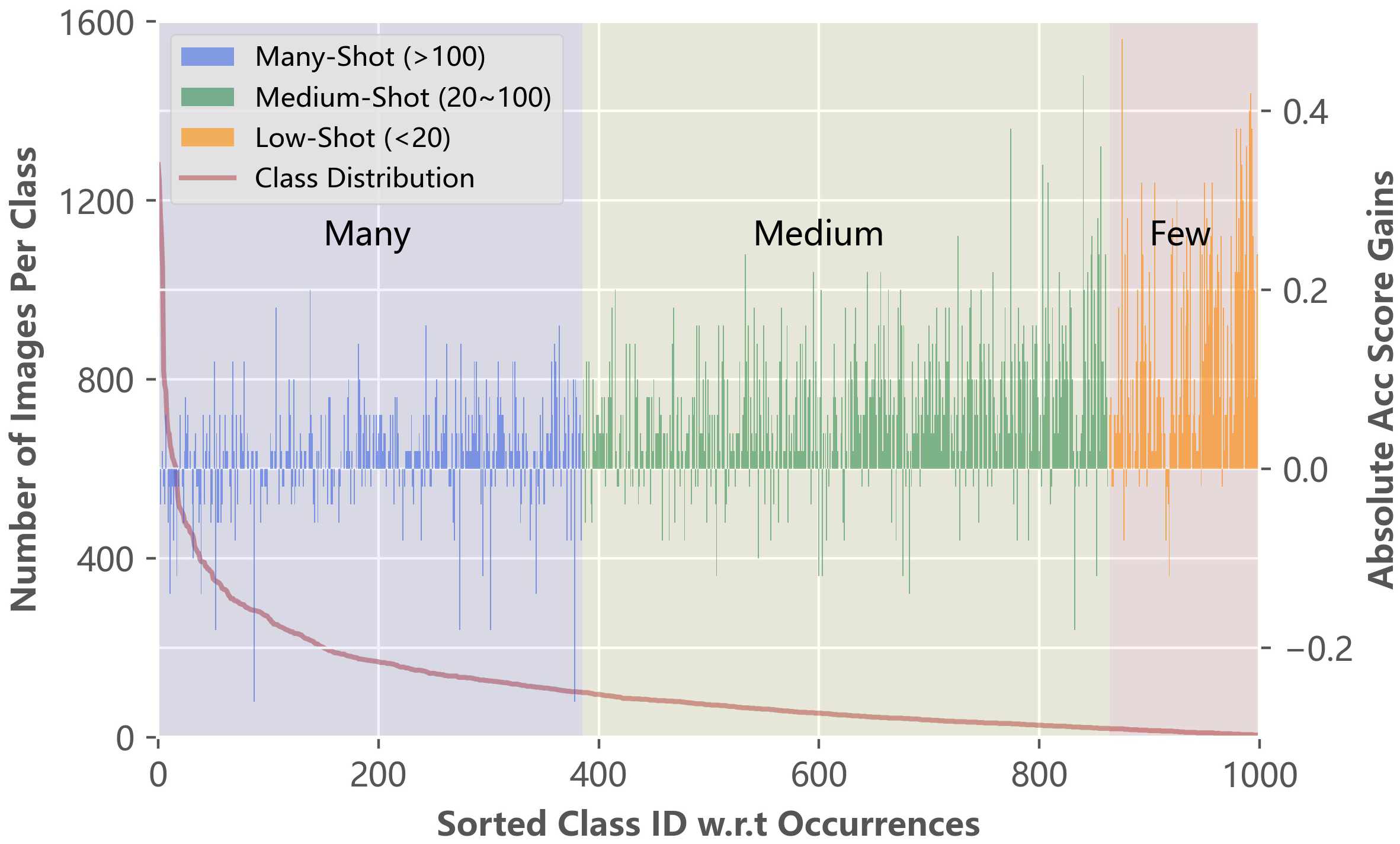}}
    	\end{center}
    	\captionsetup{font={scriptsize}}
    	% \vspace{-20pt}
    \caption{
    \textbf{Absolute accuracy score of our method over the baseline using ViT-Base/16~\cite{dosovitskiy2020image} as the backbone on ImageNet-LT~\cite{liu2019large}.}
	Our method enjoys more performance gains on classes with fewer image samples.
    }
    \label{fig:range_acc}
    % \vspace{-5pt}
\end{wrapfigure}

For a fair comparison, we also build a baseline that is only based on visual modality while keeping other settings exactly the same as our proposed method, except that the baseline models are directly initialized with CLIP pre-trained weights and fine-tuned for 100 epochs.
In addition, we re-implement and report the performance of some representative methods as well, such as $\tau$-normalized, cRT, NCM, and LWS~\cite{kang2019decoupling}, which are all initialized with CLIP pre-trained weights.

\begin{table}[t]
    \centering
    \setlength{\tabcolsep}{3mm}
    \begin{tabular}{l|l|c|c|c|c}
	\multirow{2}{*}{Method} & \multirow{2}{*}{Backbone}  & \multicolumn{4}{c}{Accuracy (\%)} \\
	\cline{3-6} 
	&  & Overall & Many & Medium & Few  \\
	\whline
	Cross Entropy \cite{li2021self} & ResNeXt-50 & 44.4 & 65.9 & 37.5 & 7.7 \\
	OLTR \cite{liu2019large} & ResNeXt-50 & 46.3 & - & - & - \\
	SSD \cite{li2021self} & ResNeXt-50 & 56.0 & 66.8 & 53.1 & 35.4 \\
	RIDE (4 Experts) \cite{wang2020long} & ResNeXt-50 & 56.8 & 68.2 & 53.8 & 36.0 \\
	TADE~\cite{zhang2021test} & ResNeXt-50 & 58.8 & 66.5 & 57.0 & 43.5 \\
	smDRAGON~\cite{samuel2021generalized} & ResNeXt-50 & 50.1 & - & - & - \\
	ResLT~\cite{cui2021reslt} & ResNeXt-101 & 55.1 & 63.3 & 53.3 & 40.3 \\
	PaCo~\cite{cui2021parametric} & ResNeXt-101 & 60.0 & 68.2 & 58.7 & 41.0 \\
	NCM \cite{kang2019decoupling} & ResNeXt-152 & 51.3 & 60.3 & 49.0 & 33.6 \\
	cRT \cite{kang2019decoupling} & ResNeXt-152 & 52.4 & 64.7 & 49.1 & 29.4 \\
	$\tau$-normalized \cite{kang2019decoupling} & ResNeXt-152 & 52.8 & 62.2 & 50.1 & 35.8 \\
	LWS \cite{kang2019decoupling} & ResNeXt-152 & 53.3 & 63.5 & 50.4 & 34.2 \\
	\hline
	NCM \cite{kang2019decoupling} & ResNet-50* & 49.2 & 58.9 & 46.6 & 31.1 \\
	cRT \cite{kang2019decoupling} & ResNet-50* & 50.8 & 63.3 & 47.2 & 27.8 \\
	$\tau$-normalized \cite{kang2019decoupling} & ResNet50* & 51.2 & 60.9 & 48.4 & 33.8 \\
	LWS \cite{kang2019decoupling} & ResNet-50* & 51.5 & 62.2 & 48.6 & 31.8 \\
	Zero-Shot CLIP \cite{clip} & ResNet-50* & 59.8   &  60.8  &  59.3  &  58.6 \\
	Baseline & ResNet-50* & 60.5 & 74.4 & 56.9 & 34.5 \\
	\rowcolor{mygray}
	VL-LTR (ours) & ResNet-50* & \bf{70.1} & \bf{77.8} & \bf{67.0} & \bf{50.8} \\
	\rowcolor{mygray}
	VL-LTR (ours) & ViT-Base* & \bf{77.2} & \bf{84.5} & \bf{74.6} & \bf{59.3} \\
	
\end{tabular}
    % \vspace{10pt}
    \caption{\textbf{Results on ImageNet-LT~\cite{liu2019large}.}
    % ``ResNet-*'' and ``ResNeXt-*'' stand for the ResNet~\cite{he2016deep} and ResNeXt~\cite{xie2017aggregated} backbone respectively.
    Our method outperforms prior arts when using a similar backbone network.
    ``*'' indicates the corresponding backbone is initialized with CLIP~\cite{clip} weights.}
    \label{tab:imnet_lt}
    % \vspace{-24pt}
\end{table}
\noindent\textbf{Results.} In Table~\ref{tab:imnet_lt}, we can see that our VL-LTR models are superior to conventional
vision-based methods with similar visual encoders (\ie, backbones).
For example, when using ResNet-50 (R-50)~\cite{he2016deep} as the backbone, the overall accuracy of our method reaches 70.1\%, which outperforms baseline by 9.6 points (70.1\% \vs 60.5\%), and 10.1 points better than previous best PaCo~\cite{cui2021parametric} (70.1\% \vs 60.0\%).

Moreover, from the aspect of few-shot accuracy, the performance of our method is more promising, which is 16.3 points and 7.3 points better than baseline (50.8\% \vs 34.5\%) and the second-best method~\cite{zhang2021test} (50.8\% \vs 43.5\%).
When replacing the backbone with heavy ViT-Base/16 (ViT-B)~\cite{dosovitskiy2020image}, the overall accuracy of our method can further boost up to 77.2\%, which is the current new state-of-the-art of ImageNet-LT, and near the prevailing performance (\ie, 80\%) training on the full ImageNet~\cite{deng2009imagenet}.

In Figure \ref{fig:range_acc}, we visualize the class-level performance improvement, which is measured by the absolute accuracy gains of our method against the baseline, both of which use ViT-B as the visual backbone. We see that there is more gains on tail classes, which indicates that our method can help mitigate the data-scarce problem under long-tail settings by introducing class-level text descriptions.

\subsection{Experiments on Places-LT}
\noindent\textbf{Settings.} We also investigate our method on Places-LT~\cite{liu2019large}, a dataset with a different domain. The experimental setting of Places-LT is the same as Section \ref{sec:imnet_lt}.

\begin{table}[t]
    \centering
    \setlength{\tabcolsep}{3mm}
    \begin{tabular}{l|l|c|c|c|c}
% 	\whline
    \renewcommand{\arraystretch}{0.1}
    %\footnotesize
	\multirow{2}{*}{Method} & \multirow{2}{*}{Backbone}  & \multicolumn{4}{c}{Accuracy (\%)} \\
	\cline{3-6} 
	&  & Overall & Many & Medium & Few  \\
	\whline
	OLTR \cite{liu2019large} & ResNet-152 & 35.9 & 44.7 & 37.0 & 25.3 \\
	ResLT~\cite{cui2021reslt} & ResNet-152 & 39.8 & 39.8 & 43.6 & 31.4 \\
	TADE~\cite{zhang2021test} & ResNet-152 & 40.9 & 40.4 & 43.2 & 36.8 \\
	PaCo~\cite{cui2021parametric} & ResNet-152 & 41.2 & 36.1 & 47.9 & 35.3 \\
	
	NCM \cite{kang2019decoupling} & ResNet-152 & 36.4 & 40.4 & 37.1 & 27.3 \\
	cRT \cite{kang2019decoupling} & ResNet-152 & 36.7 & 42.0 & 37.6 & 24.9 \\
	$\tau$-normalized \cite{kang2019decoupling} & ResNet-152 & 37.9 & 37.8 & 40.7 & 31.8 \\
	LWS \cite{kang2019decoupling} & ResNet-152 & 37.6 & 40.6 & 39.1 & 28.6 \\
	smDRAGON~\cite{samuel2021generalized} & ResNet-50 & 38.1 & - & - & - \\
	\hline
	NCM \cite{kang2019decoupling} & ResNet-50* & 30.8 & 37.1 & 30.6 & 19.9 \\
	cRT \cite{kang2019decoupling} & ResNet-50* & 30.5 & 38.5 & 29.7 & 17.6 \\
	$\tau$-normalized \cite{kang2019decoupling} & ResNet-50* & 31.0 & 34.5 & 31.4 & 23.6 \\
	LWS \cite{kang2019decoupling} & ResNet-50* & 31.3 & 36.0 & 32.1 & 20.7 \\
	Zero-Shot CLIP \cite{clip} & ResNet-50* &   38.0  &  37.5 &  37.5  &  40.1 \\
	Baseline & ResNet-50* & 39.7 & 50.8 & 38.6 & 22.7 \\
	\rowcolor{mygray}
	VL-LTR (ours) & ResNet-50* & \bf{48.0} & \bf{51.9} & \bf{47.2} & \bf{38.4} \\
	\rowcolor{mygray}
	VL-LTR (ours) & ViT-Base* & \textbf{50.1} & \textbf{54.2} & \textbf{48.5} & \textbf{42.0}  \\
	
\end{tabular}
    % \vspace{10pt}
    \caption{\textbf{Results on Places-LT~\cite{liu2019large}.}
    % ``ResNet-*'' stands for ResNet~\cite{he2016deep} backbone.
    ``*'' indicates the corresponding backbone is initialized with CLIP~\cite{clip} weights.}
    \label{tab:places_lt}
    % \vspace{-4pt}
\end{table}

\noindent\textbf{Results.} As reported in Table \ref{tab:places_lt}, using ResNet-50 (R-50) as backbone, our model achieves 48.0\% overall accuracy, surpassing counterparts by at least 6.8 points (48.0\% \vs 41.2\%), including state-of-the-art PaCo~\cite{cui2021parametric}, TADE~\cite{zhang2021test}, and ResLT~\cite{cui2021reslt}, while all of them use ResNet-152~\cite{he2016deep} as backbone. The performance are also impressive for the medium- (47.2\%) and few-shot (38.4\%) classes.
Once again, the model with ViT-Base/16 (ViT-B)~\cite{dosovitskiy2020image} gives the top overall accuracy of 50.1\%, which is a new state-of-the-art on this benchmark.

\begin{table}[t]
    \centering
    \setlength{\tabcolsep}{3.0mm}
    \begin{tabular}{l|l|c}
% 	\whline
    \renewcommand{\arraystretch}{0.1}
    %\footnotesize
	Method & Backbone  & Accuracy (\%)  \\
	\whline
	CB-Focal \cite{cao2019learning} & ResNet-50 & 61.1 \\
	LDAM+DRW \cite{cao2019learning} & ResNet-50 & 68.0 \\
	BBN \cite{zhou2020bbn} & ResNet-50 & 69.6 \\
	SSD \cite{li2021self} & ResNet-50 & 71.5 \\
	RIDE (4 experts) \cite{wang2020long} & ResNet-50 & 72.6 \\
	smDRAGON~\cite{samuel2021generalized} & ResNet-50 & 69.1 \\
	ResLT~\cite{cui2021reslt} & ResNet-50 & 72.3 \\
	TADE \cite{zhang2021test} & ResNet-50 & 72.9 \\
	PaCo \cite{cui2021parametric} & ResNet-50 & 73.2 \\
	
	NCM \cite{kang2019decoupling} & ResNet-50 & 63.1\\
	cRT \cite{kang2019decoupling} & ResNet-50 & 67.6\\
	$\tau$-normalized \cite{kang2019decoupling} & ResNet-50 & 69.3 \\
	LWS \cite{kang2019decoupling} & ResNet-50 & 69.5 \\
	\hline
	NCM \cite{kang2019decoupling} & ResNet-50* & 65.3 \\
	cRT \cite{kang2019decoupling} & ResNet-50* & 69.9 \\
	$\tau$-normalized \cite{kang2019decoupling} & ResNet-50* &  71.2 \\
	LWS \cite{kang2019decoupling} & ResNet-50* & 71.0\\
	Zero-Shot CLIP \cite{clip} & ResNet-50* &   3.4  \\
	Baseline & ResNet-50* & 72.6 \\
	\rowcolor{mygray}
	VL-LTR (ours) & ResNet-50* & \bf{74.6} \\
	\hline
	PaCo \cite{cui2021parametric} & ResNet-152 & 75.2 \\
	DeiT-B/16~\cite{touvron2020training} & - & 73.2 \\
	DeiT-B/16-384~\cite{touvron2020training} & - & 79.5 \\
	\rowcolor{mygray}
	VL-LTR (ours) & ViT-Base* & \textbf{76.8} \\
	\rowcolor{mygray}
	VL-LTR-384 (ours) & ViT-Base* & \textbf{81.0} \\
	
\end{tabular}

    % \vspace{10pt}
    \caption{\textbf{Results on iNaturalist 2018~\cite{van2018inaturalist}.}
    % ``ResNet-*'' means the ResNet~\cite{he2016deep} backbone.
    ``*'' indicates the corresponding backbone is initialized with CLIP weights.
    ``*-384'' means the input size of $384\times 384$.}
    \label{tab:inat18}
    % \vspace{-24pt}
\end{table}

\subsection{iNaturalist 2018}

\noindent\textbf{Settings.} We further test our VL-LTR on iNaturalist 2018, a long-tailed fine-grained benchmark. Following the common practice~\cite{touvron2020training},
we adopt a long training schedule. To be specific, our models are pre-trained for 100 epochs, and fine-tuned for 360 epochs.
The initial learning rate of the pre-training and fine-tuning phase is set to $5\times 10^{-4}$ and $2\times 10^{-5}$, respectively.
Correspondingly, the baseline has the same fine-tuning epochs and initial learning rate as the proposed method.
All other experimental settings are the same as Section \ref{sec:imnet_lt}.

\noindent\textbf{Results.} Table \ref{tab:inat18} shows the top-1 accuracy on iNaturalist 2018 of different methods. 
We see that when using ResNet-50 (R-50)~\cite{he2016deep} as the backbone, our models can achieve a 74.6\% overall accuracy, surpassing previous methods with the same backbone by at least 1.4 points.
Besides that, when equipped with a strong backbone ViT-Base/16 (ViT-B)~\cite{dosovitskiy2020image}, our model can have an overall accuracy of 76.8\%, which outperforms the state-of-the-art PaCo (ResNet-152) by 1.6 points (76.8\% \vs 75.2\%). 
Moreover, our model can also benefit from a larger image input size (\ie, $384\times 384$), and achieve 81.0\% top-1 accuracy, which is 1.5 points higher than DeiT-B/16-384~\cite{touvron2020training} (81.0\% \vs 79.5\%).

\begin{figure*}[t]
		\centering
		\setlength{\fboxrule}{0pt}
		\fbox{\includegraphics[width=1.0\textwidth]{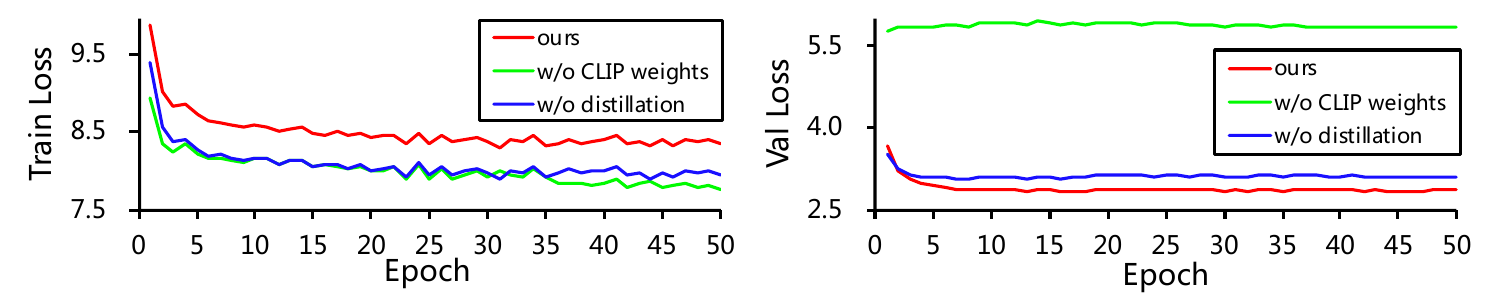}}
		% \vspace{-8pt}
		\caption{
		   		\textbf{Train and validation loss curves of VL-LTR (ResNet-50) on ImageNet-LT~\cite{liu2019large} under different settings.} Both without CLIP~\cite{clip} weights (w/o CLIP) and without distillation lead to a certain degree of overfitting.
		}
		\label{fig:loss}
		% \vspace{-16pt}
\end{figure*}

\subsection{Ablation Study}
\label{sec:abl}
\paragraph{Settings.} In order to provide a deep analysis of our proposed method, we also conduct ablation studies on the ImageNet-LT dataset.
In these experiments, we use ResNet-50 as the default backbone.
All other settings remain the same as Section \ref{sec:imnet_lt} unless specifically mentioned.

\begin{table}[t]
    \centering
    \setlength{\tabcolsep}{3.0mm}
    % \footnotesize
    \begin{tabular}{l|c|c|c|c|c|c}
    \renewcommand{\arraystretch}{0.1}
	\multirow{2}{*}{\#} & CLIP & \multicolumn{2}{c|}{Pre-training} & \multicolumn{2}{c|}{Fine-tuning} & Accuracy \\
	\cline{3-6} 
	& Weights & w/o $\mathcal{L}_\text{dis}$ & w/ $\mathcal{L}_\text{dis}$ & Head & SS & (\%) \\
	\whline
	\rowcolor{mygray}
	1 & $\checkmark$ & - & $\checkmark$ & LGR & AnSS & \textbf{70.1} \\
	2 & $\checkmark$ & - & - & LGR & AnSS & 62.8  \\
	3 & - & $\checkmark$ & - & LGR & AnSS & 46.8  \\
	4 & $\checkmark$ & $\checkmark$ & - & LGR & AnSS & 66.2  \\
	5 & $\checkmark$ & - & $\checkmark$ & FC & - & 62.1 \\
	6 & $\checkmark$ & - & $\checkmark$ & KNN & - & 63.9 \\
	7 & $\checkmark$ & - & $\checkmark$ & LGR & Cut Off &  69.7  \\
	
\end{tabular}
    % \vspace{10pt}
    \caption{
    \textbf{Ablation studies on ImageNet-LT~\cite{liu2019large}.} ``Head'' denotes the recognition head used in the fine-tuning stage, and ``SS'' denotes the sentence selection strategy.
    }
    \label{tab:abs}
    % \vspace{-20pt}
\end{table}

\noindent\textbf{Class-wise Visual-Linguistic Pre-training.} To examine the effectiveness of our class-wise visual-linguistic pre-training (CVLP) framework, we remove it by directly
performing the fine-tuning process on the pre-trained weights of CLIP~\cite{clip}. As reported in the \#1 and \#2 of Table \ref{tab:abs}, the model with CVLP outperforms the one without CVLP by 7.3 points on the overall accuracy.
Such gap might be attributed to the inconsistency between image and text representation, which can be alleviated by our CVLP.

\begin{figure}
		\centering
		\setlength{\fboxrule}{0pt}
% 		\fbox{\includegraphics[width=0.75\textwidth]{./figure/concept.pdf}}
\fbox{\includegraphics[width=1.0\textwidth]{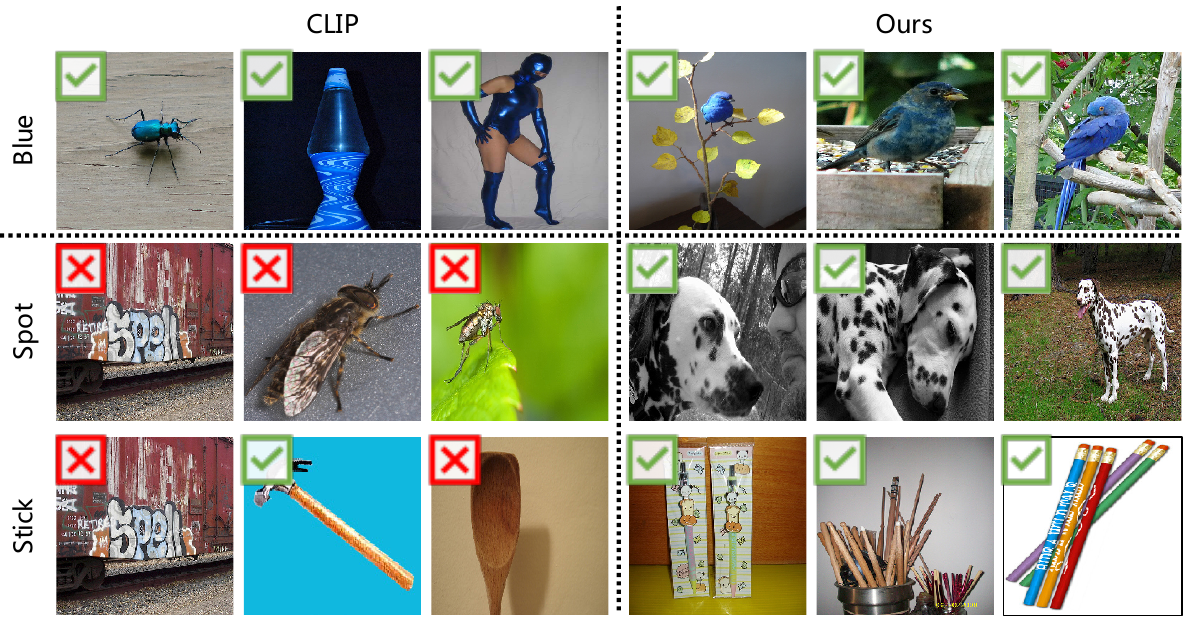}}
		% \vspace{0pt}
		\caption{
		\textbf{Concept visualization,} where ``freq'' and ``rare'' mean concepts appear frequently and rarely, respectively. Our method can effectively learn common visual concepts, and even the rare concepts where CLIP~\cite{clip} makes mistakes, such as ``spot'' texture and ``stick'' shape.
		}
		\label{fig:concept}
		% \vspace{-10pt}
\end{figure}

To verify this, we visualize some concepts by retrieving images with the greatest cosine similarity. As shown in Figure \ref{fig:concept}, both CLIP and our method can learn common visual concepts, such as the ``blue'' color, but  CLIP~\cite{clip} fails to capture rare concepts, such as ``stick'' shape and ``spot'' texture. More examples are provided in the supplementary material.

\noindent\textbf{CLIP Pre-trained Weights.} To analyze the influence of CLIP pre-trained weights, we train our method with randomly initialized weights.
Comparing the \#1 and \#3 of Table \ref{tab:abs}, we can see that initializing with CLIP pre-trained weights benefits our VL-LTR.
We also plot the curves of training and validation loss in the fine-tuning stage in Figure \ref{fig:loss}, where CLIP pre-trained weights (see red curves) can help alleviate the over-fitting problem.
This phenomenon is caused by the limited text corpus for pre-training.
There are only 1000 class descriptions (about 127K sentences) for ImageNet-LT, 
it is easy to overfit an image to a specific set of sentences
without a pre-trained linguistic encoder.

% \begin{figure}[t]
% 		\centering
% 		\setlength{\fboxrule}{0pt}
% 		\fbox{\includegraphics[width=0.44\textwidth]{./figure/loss.pdf}}
% 		\vspace{-8pt}
% 		\caption{
% 		\textbf{Train and validation loss curves of VL-LTR (ResNet-50) on ImageNet-LT~\cite{liu2019large} under different settings.} Both without CLIP~\cite{clip} weights (w/o CLIP) and without distillation lead to a certain degree of overfitting.
% 		}
% 		\label{fig:loss}
% 		\vspace{-4pt}
% \end{figure}

\noindent\textbf{Distillation Loss.} Similar to the role of pre-trained weights, the distillation loss $\mathcal{L}_\text{dis}$ is also used to reduce the risk of over-fitting in the pre-training phase.
Comparing the red and blue curves in Figure \ref{fig:loss}, the over-fitting problem is alleviated in the model with $\mathcal{L}_\text{dis}$. From the \#1 and \#4 of Table \ref{tab:abs}, we also see that the model with $\mathcal{L}_\text{dis}$ performs better than the one without $\mathcal{L}_\text{dis}$ (70.1\% \vs 66.2\%).

\noindent\textbf{Linguistic-Guided Recognition.} We verify the effectiveness of linguistic-guided recognition (LGR) by comparing it with other recognition heads, including FC (vision-based), and KNN (vision-language-based).
As reported in \#1, \#5, and \#6 of Table \ref{tab:abs}, the proposed LGR performs better than FC and KNN by 8.0\% and 6.2\% points in overall accuracy respectively.
It is notable that, as a vision-language-based recognition head, KNN also works better than FC.
These results indicate the effectiveness of LGR and the power of visual-linguistic representation.

\noindent\textbf{Anchor Sentence Selection.} We study the effectiveness of anchor sentence selection (AnSS) by replacing it with ``Cut Off'' strategy, 
where we simply select the first $M$ sentences from text descriptions as the anchor sentences for visual recognition.
As shown in Table \ref{tab:abs}, the model with AnSS (see the \#1 of Table \ref{tab:abs}) outperforms the model with ``Cut Off'' on the overall accuracy,
% We also see in Figure * that, AnSS can filter out some noisy sentences.
which proves the effectiveness of AnSS to filter out some noisy sentences.
Note that, AnSS is a training-free module, which can bring considerable improvements in noisy scenes.

% \begin{table}[t]
%     % \vspace{10pt}
%     \centering
%     % \renewcommand\arraystretch{0.97}
%     \setlength{\tabcolsep}{3mm}
%     % \input{table/speed.tex}
%     \input{table/speed1.tex}
%     % \vspace{10pt}
%     \caption{\textbf{Computation overhead comparison of our VL-LTR (ResNet-50) and the baseline (ResNet-50).} Our method has almost the same GFLOPs and inference speed to the baseline. GFLOPs is calculated under the input scale of $224\times 224$.}
%     \label{tab:speed}
%     % \vspace{-4pt}
% \end{table}

% \subsection{Computation Overhead}

% As mentioned in Section \ref{sec:overall}, our VL-LTR is a two-stage framework with two encoders.
% %
% Nevertheless, we would like to point out that the computational cost of our method is almost the same as the vision-based method, 
% %
% since the linguistic encoder is not necessary at the inference stage.
% %
% Specifically, after pre-training, the text embeddings of anchor sentences can be pre-populated offline.
% %
% During inference, we only need to load the pre-populated text embeddings to perform visual recognition.
% %
% As reported in Table \ref{tab:speed}, the GFLOPs and the inference speed of our method are similar to the baseline.
% %
% These results are tested with a batch size of 128 on one V100 GPU and one 2.20GHz CPU in a single thread.

\subsection{Limitations}
Although the proposed VL-LTR achieves good performance on multiple long-tailed recognition benchmarks, it still has some flaws.
%
% First, due to the limited text corpus, our method heavily relies on pre-trained weights to learn high-quality linguistic representation.
% Second, like most LTR works~\cite{li2021self,cui2021parametric,kang2019decoupling}, our VL-LTR is a two-stage method as well, which cannot be end-to-end trained.
% But we believe these problems could be well addressed with the development of data collection and visual-linguistic model in the future.
First, due to the limited text corpus, our method currently relies on existing pre-trained foundation models to learn high-quality linguistic representation.
Second, like most LTR works~\cite{li2021self,cui2021parametric,kang2019decoupling}, our VL-LTR is a two-stage method as well, which does not support end-to-end training.
But we believe these problems could be well addressed in the future with the enrichment of text data and the development of visual-linguistic model.

\section{Conclusions}

In this work, we introduce VL-LTR, a new visual-linguistic framework for long-tailed recognition. 
We develop a class-level visual-linguistic pre-training (CVLP) to connect images and text descriptions at class level, and a language-guided recognition (LGR) head to make effective use of visual-linguistic representation for visual recognition.
Extensive experiments on various long-tailed recognition benchmarks verify that our method works better than well-designed vision-based methods.
We hope this work could provide a strong baseline for vision-language-based long-tailed visual recognition.

% \clearpage
\bibliographystyle{splncs04}
\bibliography{egbib}

% \newpage
\section{Appendices}

% \begin{appendices}

\appendix
\section{Methodology Details}
\label{app:met}
For convenience, we summarize all the notations used in the paper in Table~\ref{tab:notations}.

\begin{table}[h]
    \centering
    \setlength{\tabcolsep}{3mm}
    \caption{\textbf{Summary of notations used in the paper.}}
    \begin{tabular}{l|l}
    \renewcommand{\arraystretch}{0.1}
    Notation & Meaning \\
    \whline
    $\mathcal{I}=\left\{I_i \right\}_{i=1}^N$ & A batch of $N$ image samples \\
    $\mathcal{T}=\left\{T_i \right\}_{i=1}^N$ & A batch of $N$ text samples  \\
    $M$ & Number of anchor sentences per class \\
    $\mathcal{E}_\text{vis}(\cdot)$ &  Visual encoder\\
    $\mathcal{E}_\text{lin}(\cdot)$ &  Linguistic encoder\\
    $E^I_i$ &  Embeddings of image $I_i$\\
    $E^T_i$ &  Embeddings of text $T_i$\\
    $S_{i,j}$ & Cosine similarity of $E^I_i$ and $E^T_j$\\
    $\left<E^I, G\right>$ & Cosine similarity of $E^I$ and $G$ \\
    $\mathcal{L}_\text{ccl}$ & Class-wise contrastive loss \\
    $\mathcal{L}_\text{vis}$ & Class-wise contrastive loss for images \\
    $\mathcal{L}_\text{lin}$ & Class-wise contrastive loss for texts \\
    $\mathcal{L}_\text{dis}$ & Distillation loss \\
    $\mathcal{L}_\text{pre}$ & Pre-training loss \\
    $\mathcal{L}_\text{rec}$ & Recognition loss \\
    $\mathbf{y}$ & Ground truth label \\
    
\end{tabular}
    \label{tab:notations}
\end{table}

\section{Class-level Corpus Preparation}
\label{app:exp}

\begin{figure*}[htb]
		\centering
		\setlength{\fboxrule}{0pt}
		\fbox{\includegraphics[width=1.0\textwidth]{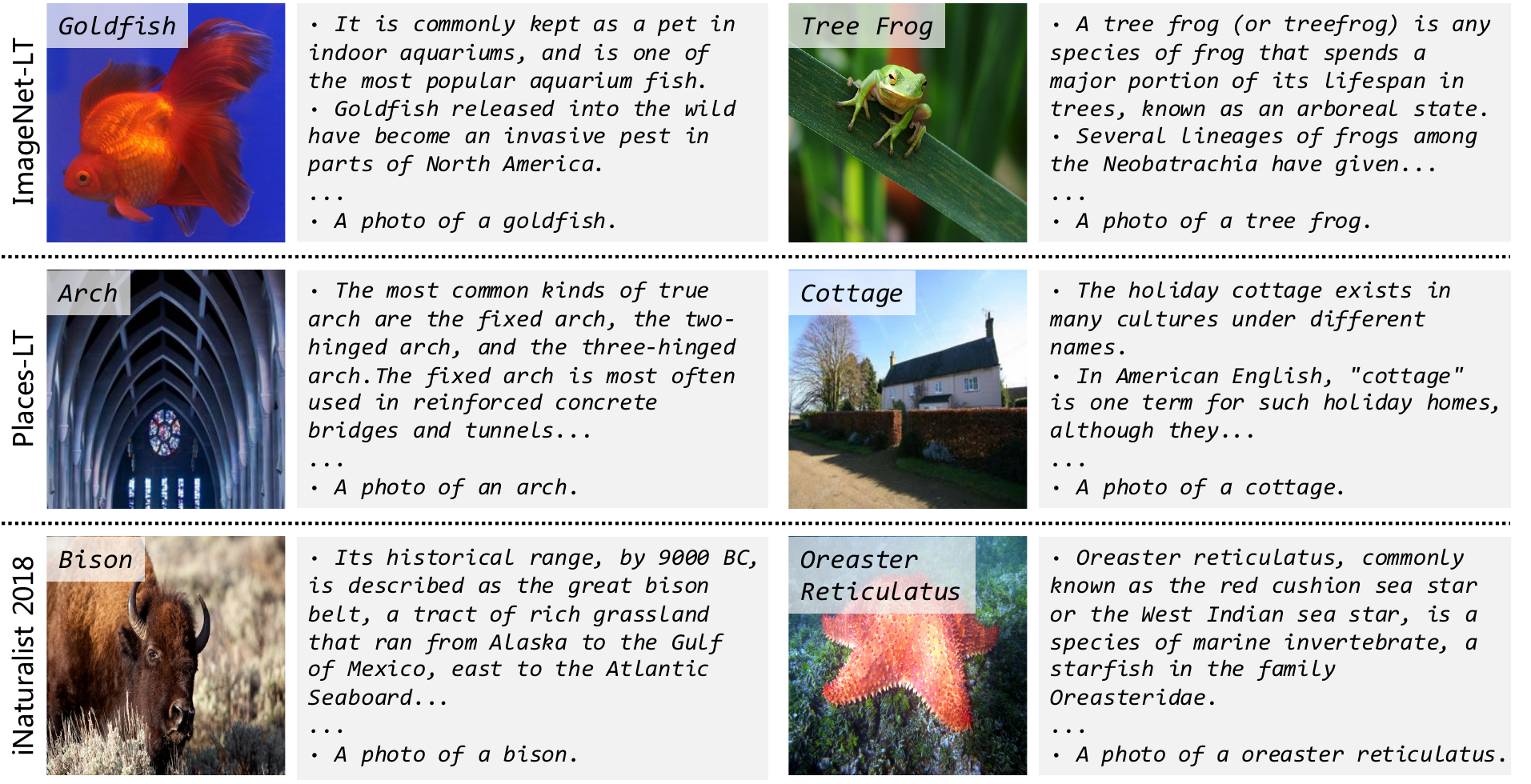}}
		\caption{
		\textbf{Examples of text descriptions crawled from Wikipedia for ImageNet-LT~\cite{liu2019large}, Places-LT~\cite{liu2019large} and iNaturalist-2018~\cite{van2018inaturalist}}, in which both redundant useful and noise information can be found.
		}
		\label{fig:text}
\end{figure*}

As described in Section 4.1, we collect class-level text descriptions from Wikipedia and prompt templates provided in \cite{clip}.
In Figure~\ref{fig:text}, we display part of text descriptions collected for ImageNet-LT~\cite{liu2019large}, Places-LT~\cite{liu2019large}, and iNaturalist-2018~\cite{van2018inaturalist} datasets. We see that since these texts are all crawled from the Internet, it is inevitable to have some noisy text within them.

\begin{table}[h]
    \centering
    \setlength{\tabcolsep}{3mm}
    \caption{\textbf{Detailed statistics of the class-level text descriptions for each dataset}, where $M_\text{min}$, $M_\text{max}$, $M_\text{mean}$, and $M_\text{Med}$ denotes the minimum, maximum, mean, and median number of sentences of classes respectively, and $L_\text{Avg}$ denotes the average number of tokens per sentence.}
    \begin{tabular}{l|c|c|c|c|c}

Dataset          & $M_\text{min}$ & $M_\text{max}$ & $M_\text{mean}$ & $M_\text{Med}$ & $L_\text{Avg}$ \\ \whline
ImageNet-LT~\cite{liu2019large}     &   1  &  721   &   127   &   89  &  29          \\
Places-LT~\cite{liu2019large}        &   2  &  610   &    116  &  77   &  29          \\
iNaturalist 2018~\cite{van2018inaturalist} &   1  &  1774  &   33    &  17   &  26          \\ 
\end{tabular}
    \label{tab:text_stats}
\end{table}

In addition, we report detailed statistics of the collected text descriptions in Table~\ref{tab:text_stats}, where we find that even if all the corpus comes from Wikipedia, the text quantity of different classes varies greatly.

\section{Computation Overhead}

As mentioned in Section 3.1, our VL-LTR is a two-stage framework with two encoders.
Nevertheless, we would like to point out that the computational cost of our method is almost the same as the vision-based method, 
since the linguistic encoder is not necessary at the inference stage.
Specifically, after pre-training, the text embeddings of anchor sentences can be pre-populated offline.
During inference, we only need to load the pre-populated text embeddings to perform visual recognition.
As reported in Table \ref{tab:speed}, the GFLOPs and the inference speed of our method are similar to the baseline.
These results are tested with a batch size of 128 on one V100 GPU and one 2.20GHz CPU in a single thread. 
Moreover, we believe such conclusion also applies to other backbones such as ViT, Swin, TransFG, and complemental attention, since our framework is orthogonal to the backbone's structure.

\begin{table}[h]
    % \vspace{10pt}
    \centering
    \setlength{\tabcolsep}{3mm}
    \caption{\textbf{Computation overhead comparison of our VL-LTR (ResNet-50) and the baseline (ResNet-50).} Our method has almost the same GFLOPs and inference speed to the baseline. GFLOPs is calculated under the input scale of $224\times 224$.}
    \begin{tabular}{l|c|c}
    \renewcommand{\arraystretch}{0.1}
	Method & GFLOPs & Time Cost (ms) \\
	\whline
	Baseline & 5.4 & 1.1 \\
	\rowcolor{mygray}
	VL-LTR (ours) & 5.5 & 1.3 \\
	
\end{tabular}
    % \vspace{10pt}
    \label{tab:speed}
    % \vspace{-4pt}
\end{table}

\section{Comparison with Zero-Shot CLIP}

In Table~\ref{tab:zero_shot}, we compare our results and the zero-shot results of CLIP~\cite{clip} on ImageNet-LT~\cite{liu2019large}, Places-LT~\cite{liu2019large} and iNaturalist 2018~\cite{van2018inaturalist} datasets, respectively. We see that the performance of CLIP drops sharply when the domain of target data (\eg, iNaturalist 2018) is inconsistent with its training data,
while our method can achieve significant improvement on all datasets.

\begin{table}[t]
    \centering
    \setlength{\tabcolsep}{3mm}
    \caption{
    \textbf{Comparison with Zero-Shot CLIP}. Our method achieves improvements on all datasets and is robust to datasets of different domains.
    }
    \begin{tabular}{l|c|c|c|c|c}
\multirow{2}{*}{Dataset}          & \multicolumn{1}{c|}{\multirow{2}{*}{Method}} & \multicolumn{4}{c}{Accuracy(\%)} \\
\cline{3-6} 
    & \multicolumn{1}{c|}{}
    & Overall  & Many  & Medium  & Few \\ \whline
\multirow{3}{*}{\tabincell{l}{ImageNet-LT}}      & Zero-Shot                           
    &   59.8   &  60.8  &  59.3  &  \textbf{58.6} \\
                                  & Baseline
    &   60.5   &  74.4  &  56.9  &  34.5 \\
                                  %\rowcolor{mygray}
                                  & \cellcolor{mygray}VL-LTR (ours)
    & \cellcolor{mygray}\textbf{70.1}   & \cellcolor{mygray}\textbf{77.8}  &  \cellcolor{mygray}\textbf{67.0}  &  \cellcolor{mygray}50.8 \\ \hline
\multirow{3}{*}{\tabincell{l}{Places-LT}}        & Zero-Shot                                  &   38.0    &  37.5 &   37.5  &  \textbf{40.1} \\
                                  & Baseline
    &   39.7    &  50.8   &  38.6   &  22.7   \\
                                  %\rowcolor{mygray}
                                  & \cellcolor{mygray}VL-LTR (ours)
    &  \cellcolor{mygray}\textbf{48.0}   &  \cellcolor{mygray}\textbf{51.9} &  \cellcolor{mygray}\textbf{47.2}  &  \cellcolor{mygray}38.4  \\ \hline
\multirow{3}{*}{\tabincell{l}{iNaturalist 2018}} & Zero-Shot                                  &    3.4      &   6.1   &   3.3   &  2.9 \\
                                  & Baseline
    &   72.6   &  76.6  &   74.1  &  70.2 \\
                                  %\rowcolor{mygray}
                                  & \cellcolor{mygray}VL-LTR (ours)
    &   \cellcolor{mygray}\textbf{74.6}       &  \cellcolor{mygray}\textbf{78.3}  &  \cellcolor{mygray}\textbf{75.5}   &   \cellcolor{mygray}\textbf{72.7}
\end{tabular}
    \label{tab:zero_shot}
\end{table}

% \vspace{-0.35cm}
\section{Comparison of Different Distillation Methods in CVLP}
To further study the influence of distillation in the pre-training phase, we try to use the pre-trained CLIP model~\cite{clip} as the teacher model to distill the visual and linguistic encoder of our model at the feature level, in addition to the logits distillation mentioned in Section 3.2.
As reported in Table~\ref{tab:distill}, both feature distillation and logits distillation can improve recognition accuracy, and our method achieves the highest accuarcy on ImageNet-LT~\cite{liu2019large} when using logits distillation with the loss weight $\lambda$ of 0.5.
\begin{table}[h]
    \centering
    \setlength{\tabcolsep}{3mm}
    \caption{
    \textbf{Results of different types of distillation in CVLP
    on ImageNet-LT~\cite{liu2019large}.} Our method achieves the highest accuarcy when using logits distillation with the loss weight $\lambda$ of 0.5.
    }
    \begin{tabular}{l|c|c|c|c|c}
\multirow{2}{*}{Distill Level} & \multirow{2}{*}{$\lambda$} & \multicolumn{4}{c}{Accuracy (\%)} \\
\cline{3-6}
&  & Overall  & Many  & Medium  & Few  \\
\whline
- & 0 & 66.2  & 76.9 & 63.5 & 42.5 \\
\hline
\multirow{2}{*}{Feature}    
& 0.1 &    67.3      &   77.3    &   64.4      &  44.0    \\
& 0.5 &     68.0     &    77.6   &    65.2     &   45.5   \\
\hline
\multirow{2}{*}{Logits}        
& 0.1 &    68.3      &   \textbf{77.9}    &    65.3     &  45.1    \\
& \cellcolor{mygray}0.5 (ours)   
& \cellcolor{mygray}\textbf{70.1}
& \cellcolor{mygray}77.8
& \cellcolor{mygray}\textbf{67.0}
& \cellcolor{mygray}\textbf{50.8}\\
\end{tabular}
    % \vspace{4pt}
    \label{tab:distill}
\end{table}

\begin{table*}[t]
    \centering
    \setlength{\tabcolsep}{3mm}
    \caption{\textbf{Results of using different text source on ImageNet-LT~\cite{liu2019large} and Places-LT~\cite{liu2019large},} where we see that
    ``wiki + prompt'' outperforms ``prompt only'' in overall, medium, and few accuracy.
    }
    \begin{tabular}{l|c|c|c|c|c}
\multirow{2}{*}{Dataset}          & \multicolumn{1}{c|}{\multirow{2}{*}{Source}} & \multicolumn{4}{c}{Accuracy(\%)} \\
\cline{3-6} 
    & \multicolumn{1}{c|}{}
    & Overall  & Many  & Medium  & Few \\ \whline
\multirow{3}{*}{ImageNet-LT}
& Baseline
& 60.5 & 74.4 & 56.9 & 34.5 \\
& prompt only                          
&   69.4   &  \textbf{77.9}  &  66.5  &  49.3 \\
& \cellcolor{mygray}wiki + prompt (ours)
&   \cellcolor{mygray}\textbf{70.1}   &  \cellcolor{mygray}77.8  &  \cellcolor{mygray}\textbf{67.0}  &  \cellcolor{mygray}\textbf{50.8} \\ \hline
\multirow{3}{*}{Places-LT}
& Baseline
& 39.7 & 50.8 & 38.6 & 22.7 \\
& prompt only                                &   47.3    &  \textbf{52.7} &   46.8  &  36.3 \\
%\rowcolor{mygray}
& \cellcolor{mygray}wiki + prompt (ours)
    &  \cellcolor{mygray}\textbf{48.0}   &   \cellcolor{mygray}51.9 &  \cellcolor{mygray}\textbf{47.2}  &  \cellcolor{mygray}\textbf{38.4}
\end{tabular}
    % \vspace{4pt}
    \label{tab:text_source}
    % \vspace{-4pt}
\end{table*}

\section{Comparison of Different Text Description Sources}
In Table~\ref{tab:text_source}, we compare the results of models using different kinds of text descriptions on ImageNet-LT~\cite{liu2019large}. 
Specifically, we use the prompt sentences provided in~\cite{clip} as the source of text description. We mark this model as ``prompt only'', and compare it with the default model that uses both Wikipedia and prompt templates as the source of text description (\ie, ``wiki + prompt'').
We see that ``wiki + prompt'' outperforms ``prompt only'' in overall, medium, and few accuracy, which demonstrates the effectiveness of corpus from Wikipedia.

We also notice that although ``prompt only'' is not the best, its performance is still relatively competitive compared to the vision-based methods (\eg, the strong Baseline established in this work).
We attribute this phenomenon to reasons as follows: (1) Our method can make effective use of the pre-trained image and text encoder of CLIP~\cite{clip}, while vision-based methods can only use image encoder;
(2) Some class names themselves contain discriminative language information, such as ``gold fish'', ``tree frog'', and ``mountain bike''.

\section{Visualization of AnSS}

\begin{figure*}[t]
		\centering
		\setlength{\fboxrule}{0pt}
		\fbox{\includegraphics[width=1.0\textwidth]{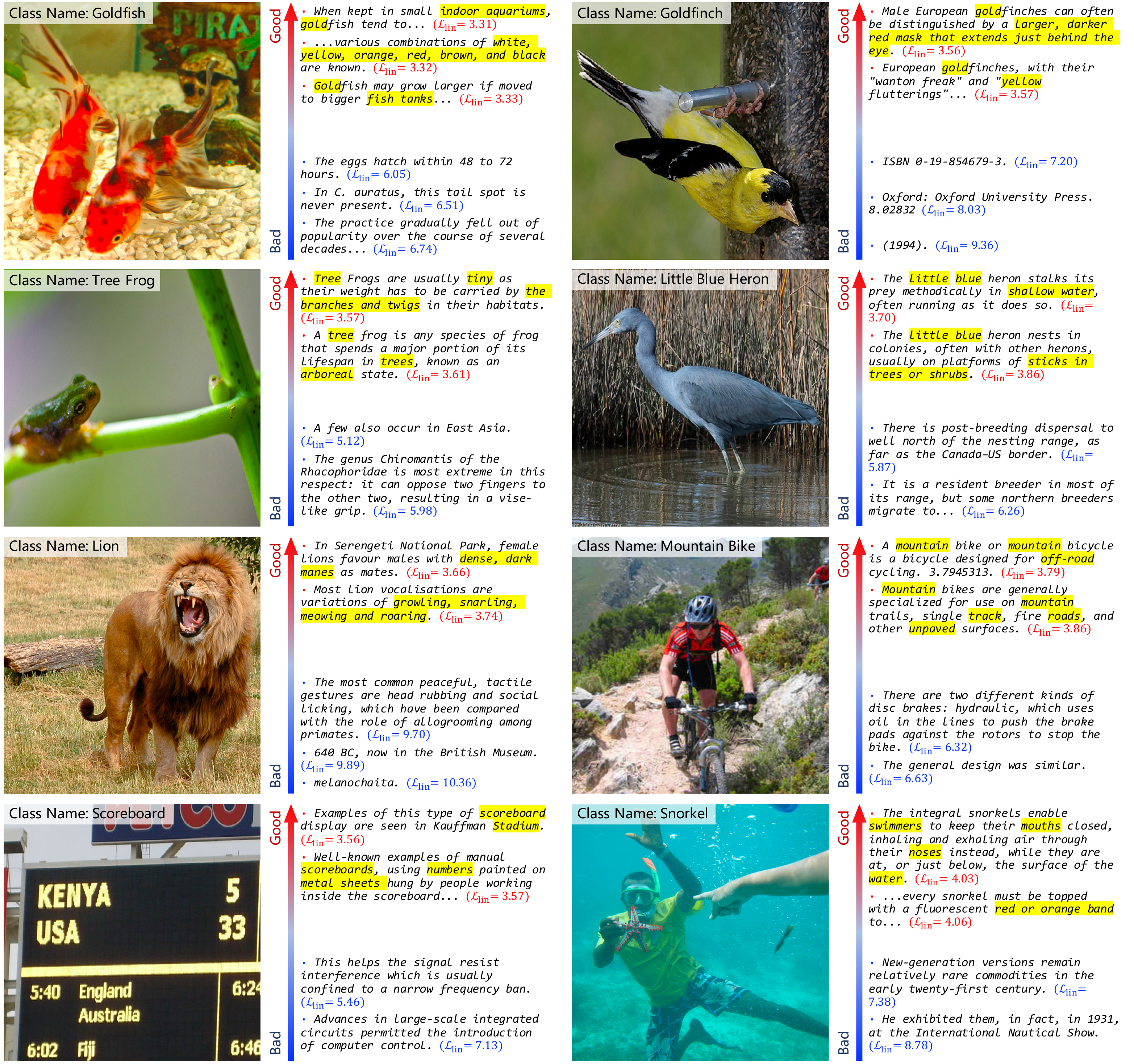}}
		\caption{
		\textbf{Some ``good'' and ``bad'' sentences and their corresponding $\mathcal{L}_{\text{lin}}$ of classes in ImageNet-LT~\cite{liu2019large}.}
		The value of $\mathcal{L}_{\text{lin}}$ can reflect the usefulness of these sentences to some extent, which thereby supports the effectiveness of our AnSS.
		}
		\label{fig:tce}
\end{figure*}

To intuitively show the effectiveness of our anchor sentence selection (AnSS), we also present some sentences recommended or filtered out by our AnSS of different classes in Figure~\ref{fig:tce}. We see that our method can reserve useful texts and drop the useless ones effectively.

\section{More Examples of Concept Visualization}

\begin{figure*}[t]
        % \vspace{-0.8cm}
		\centering
		\setlength{\fboxrule}{0pt}
		\fbox{\includegraphics[width=1.0\textwidth]{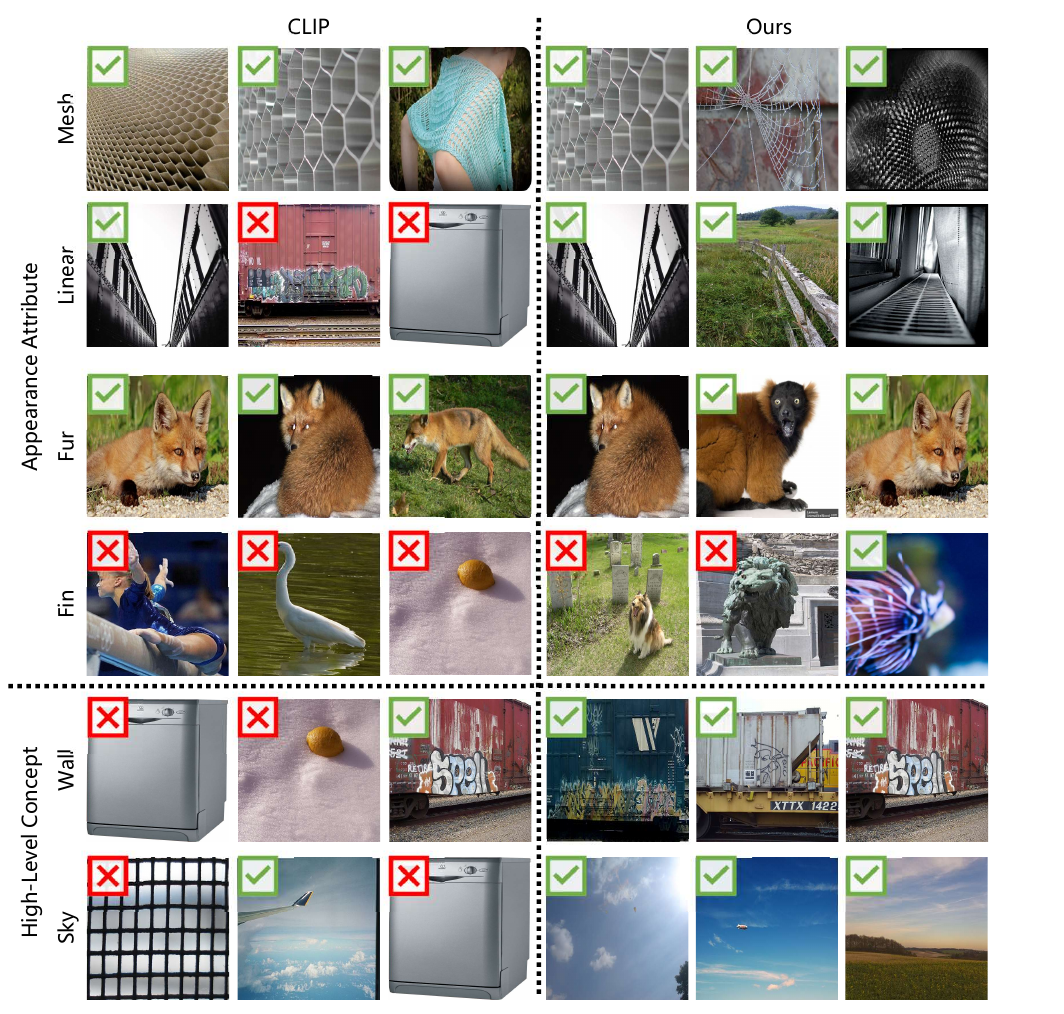}}
		% \vspace{-5pt}
		\caption{
		\textbf{Examples of concept visualization.} Our method can not only learn the texture (\eg, mesh) and shape (\eg, linear) of objects, but can also understand some visual attributes (\eg, fur and fin) and high-level concepts (\eg, wall and sky).
		In addition, compared to the original CLIP~\cite{clip}, our method can cover more visual concepts.
		}
		\label{fig:vis}
		% \vspace{-10pt}
\end{figure*}

In this section, we provide more concept visualization results of VL-LTR (ResNet-50) trained on ImageNet-LT~\cite{liu2019large}. As shown in Figure~\ref{fig:vis}, our models can not only learn some appearance attributes such as the shape and texture, but also understand high-level concepts like ``wall'' and ``sky''. 
Moreover, benefiting from CVLP, our method can cover more visual concepts than CLIP.

% \bibliography{egbib}

\end{document}